\documentclass[
]{ceurart}

\usepackage{listings}
\usepackage{subcaption}
\begin{document}

\copyrightyear{2022}
\copyrightclause{Copyright for this paper by its authors.
  Use permitted under Creative Commons License Attribution 4.0
  International (CC BY 4.0).}

\conference{Woodstock'22: Symposium on the irreproducible science,
  June 07--11, 2022, Woodstock, NY}

\title{Improving Similar Case Retrieval Ranking
Performance from Learning-to-Rank Perspective}

\tnotemark[1]
\tnotetext[1]{You can use this document as the template for preparing your
  publication. We recommend using the latest version of the ceurart style.}

\author[1]{Yuqi Liu}[%
orcid=0009-0009-6526-7658,
email=211224027@cupl.edu.cn,
url=https://liuyuqi123study.github.io,
]
\cormark[1]

\author[1]{Yan Zheng}[%
orcid=0000-0001-7116-9338,
email=zhengyan@cupl.edu.cn,
]

\author[1]{Guo Zhou}[%
email=guo.zhou@live.com,
]

\address[1]{China University of Political Science and Law, 27 Fuxue Road, Beijing, China}

\cortext[1]{Corresponding author.}

\begin{abstract}
  Given the rapid development of Legal AI, a lot of attention has been paid to
one of the most important legal AI tasks – similar case retrieval, especially with
language models to use. In our paper, however, we try to improve the ranking
performance of current models from the perspective of learning to rank instead of
language models. Specifically, we conduct experiments using a pairwise method – 
RankSVM as the classifier to substitute a fully connected layer, combined with
commonly used language models on similar case retrieval datasets LeCaRDv1
and LeCaRDv2. We finally come to the conclusion that RankSVM could generally
help improve the retrieval performance on the LeCaRDv1 and LeCaRDv2
datasets compared with original classifiers by optimizing the precise ranking. It
could also help mitigate overfitting owing to class imbalance. Our code is available
in https://github.com/liuyuqi123study/RankSVM for SLR
\end{abstract}


\maketitle

\section{Introduction}
In recent years, Legal Artificial Intelligence (AI) has attracted attention from both
AI researchers and legal professionals\cite{greenleaf2018building}. Legal AI mainly means
applying artificial intelligence technology to help with legal tasks. Benefiting from the
rapid development of AI, especially within the natural language processing (NLP) domain,
Legal AI has had a lot of achievements in real-world law applications\cite{zhong-etal-2020-nlp,surden2019artificial} such as judgment document generation, legal elements extraction, Legal AI inquiries, and judgment prediction, etc..
Among legal AI tasks, similar case retrieval (SCR) is a representative legal AI
application, as the appeal to similar sentences for similar cases plays a pivotal role in
promoting judicial fairness and may benefit other applications using large language models(LLM) under
settings like RAG\cite{wiratunga2024cbr}.
As demonstrated in a lot of work, there are mainly two approaches to do information
retrieval. 1) Traditional IR models(e.g.BM25 which is a probabilistic retrieval
model\cite{BM25} based on frequency). 2) More advanced techniques using pre-trained models with deep learning skills. The former method tends to overlook the lexical information within texts as sometimes different words may have the same semantic meaning. The latter, mainly based on transformers\cite{DBLP:journals/corr/abs-1708-02002} can achieve
promising results\cite{zhong-etal-2020-nlp}, which is also the focus of our work. According to this overview, within the similar case retrieval task or even the legal artificial intelligence, the improvements could be classified into two directions -- research based on embeddings and research based on symbols. The first kind focuses on the optimization of word embeddings as many researchers propose that the representation given by pretrained models lacks consideration of legal knowledge within legal texts. The other stresses the extraction of legal relations and events within legal texts, as it helps to improve the interpretability of legal artificial intelligence.
When using pre-trained language models, there have been a lot of variations endeavoring in both ways, trying to incorporate
the structural information and legal elements in legal documents to help with the retrieval. In BERT\_LF, for example, based on representation given by BERT, it also integrates topic information and entity information\cite{hu2022bert_lf}. In BERT\_PLI\cite{Shao2020BERTPLIMP}, it designs a model to capture the semantic relationship between paragraphs. In Ma et al.'s work\cite{10.1145/3609796}, they point out that we need to consider the internal structural information in legal texts especially legal judgments, so they respectively encode the fact part, reasoning part, and sentence part of legal judgments. There is also some work using knowledge graphs instead of language models to encode legal texts, emphasizing the encoding of events to interpretability of the model\cite{ZHANG2024103729}. Some people have also proposed to utilize LLMs to boost similar case
retrieval\cite{zhou2023boostinglegalcaseretrieval}.
In our paper, we instead do not focus on the optimization of language model architecture, but try to optimize classifiers used in the task from a learning-to-rank perspective. Even though the work on the model architecture so far is significant, we want to improve the ranking performance from the problem kind itself. As it is pointed out in early work, there are two important factors to take into account in document retrieval -- to rank the most relevant cases more correctly and to consider that different queries may have different numbers of queries\cite{10.1145/1148170.1148205}. As a problem of ranking
by criterion of relevance, however, SCR is reduced to a 2-class classification problem
following a pointwise path in a lot of work, which means the related classifier is only
used to produce one label for a query-candidate pair without actual comparison. In our work, we are aimed at solving two problems – first to  have higher accuracy on top-ranked
documents and second to consider each query case by case, in the hope of mitigating class imbalance. 
In our paper, we reintroduce pairwise methods to do the final ranking. Under
pairwise settings, classifiers care about the relative order between two documents,
which is closer to the actual ranking task\cite{liu2009learning}. Among pairwise approaches,
we pay extra attention to the Ranking Support Vector Machine(RankSVM) method
as a representative method. The advantage of RankSVM is that it aims to minimize
the ranking loss and can also mitigate the negative influence of the class-imbalance
issue\cite{10.5555/2981345.2981385,DBLP:journals/corr/WuZ16} which is common for SCR
tasks. 

\begin{figure}
  \centering
  \includegraphics[width=\linewidth]{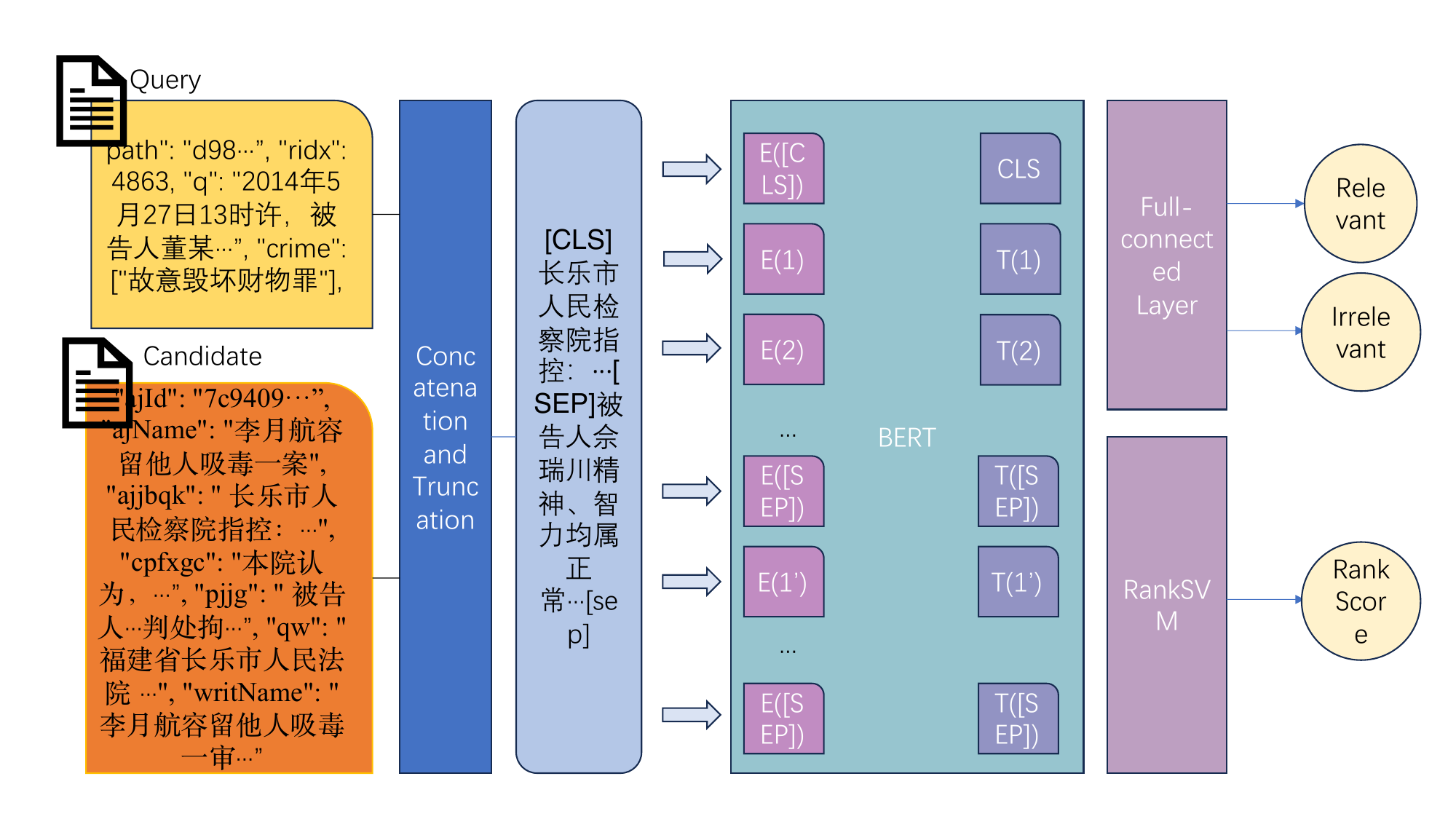}
  \caption{Illustraction of our Method}
  \label{GA}
\end{figure}

In our experiments, we test different kinds of retrieval models when combined with
RankSVM to examine their retrieval performance on different similar case retrieval datasets LeCaRDv1\cite{ma2021lecard} and LeCardv2\cite{li2023lecardv2largescalechineselegal}. Our method is
illustrated in Figure\ref{GA}. 

Our contribution could be summarized as follows, 
\begin{itemize}
    \item We first compare different kinds of classifiers and ranking functions used for the final step ranking and give explanations.
    \item We first consider SSM architecture in similar case retrieval tasks and give comparisons.
    \item We give an efficient practice of how pairwise methods could help improve the ranking performance in SCR task.
    
\end{itemize}
\section{Related Work}
\subsection{Learning-to-rank and Ranking Functions}
Learning-to-rank is a long-studied and fruitful field, and people have started thinking about it in similar case retrieval tasks. While it is a common practice to use cross-entropy and fully-connected classifiers for SCR tasks, many researchers have explored different classifiers. For example, it
is also mentioned by other papers that\cite{10.1007/978-3-031-10986-7_43} the fine-tuned BERT model
does not compare which case candidate is more similar to the query case, and they also move on to use a pairwise method borrowing its idea from RankNet\cite{10.1145/1102351.1102363} on the dataset LeCaRDv1, where the optimization goal is also the number of correctly compared pairs. The difference between RankNet and their method is that they give a different quantity of probability. At the same time, some people also employ InfoNCE in retrieval tasks\cite{xu2024rankmambabenchmarkingmambasdocument}, which is a popular method under a self-supervised setting. Though it considers both positive data and negative data in its optimization goal, it overlooks the relative relationship among different samples. At the same time, as it needs one positive sample during each sampling and always encodes a whole sampling group, it may take a lot of memory and computation compared with pairwise settings. In Ma's work\cite{10.1145/3609796}, they also discard the two-class classifier to use LightGBM\cite{10.5555/3294996.3295074}, which originates from gradient boosting decision trees. This is also a pointwise method optimizing the final ranking score. SVM, however, is a relatively old method, but it is still in use in document retrieval. For example, in work by Vacek et al., they integrate SVMs with CNN and GRU in document retrieval \cite{vacek-etal-2019-litigation-analytics}. So far, there lacks comprehensive comparison over pairwise method and pointwise method particularly fully-connected classifier for SCR tasks.
\subsection{RankSVM}
RankSVM, as it is said before, is a representative pairwise method. The framework of the algorithm itself was proposed by Joachims in 1998\cite{Joachims1998MakingLS}, with the optimization problem it's aimed at solving built in 2002\cite{10.1145/775047.775067}. It was thought to be able to optimize the retrieval quality with clickthrough data. In search engines, the data could be written in the form of $(q,r,c)$, $q$ is the query, $r$ is the ranking, and $c$ is the clickthrough data finally chosen by the user. For this kind of information retrieval system, it could be formalized as follows. For a query $q$ and a document set $D=\{d_1,\dots,d_m\}$, the best retrieval system would return a ranking $r^*$, which corresponds to the relevance between documents in $D$ and query $q$. The evaluation of how good a retrieval function is lies in how its ranking $r_{f(q)}$ can estimate the optimal value. Here $r^*$ and $r_{f(q)}$ are both binary relations defined on $D\times D$ and satisfy properties of a weak ordering , which means that $r^*\subset D\times D$ and $r_{f(q)} \subset D \times D$ being asymmetric, and negatively transitive. If a document $d_i$ is ranked higher than $d_j$ for an ordering $r$,i.e. $d_i\textless_r d_j$ then $(d_i,d_j)\in r$, otherwise $(d_i,d_j) \in r$.  In this early paper, it has also talked about the measure of similarity between the system ranking $r_{f(q)}$ and the target ranking $r^*$. While average precision is a popular choice, it should be ditched for reasons similar to our view that binary relevance is very coarse. It decides to use Kendall's $\tau$ as the parameter to be optimized. For $r_a \subset D\times D$ and $r_b \subset D\times D$, the calculation of Kendall's $\tau$ is based on the number $P$ of concordant pairs and the number $Q$ of discordant pairs and could be stated as follows
\begin{equation}
    \tau(r_a,r_b)=\frac{P-Q}{P+Q}=1-\frac{2Q}{\binom{m}{2}}
\end{equation}
Under the setting of binary classification evaluation, maximizing this quantity is minimizing the average ranking of relevant documents. Then, the goal of learning a ranking function is to find a retrieval function $f(q)$, so that the $\tau$ would satisfy
\begin{equation}
\tau_p(f)=\int \tau(r_{f(q)},r^*)dPr(q,r^*)
\end{equation}
But a lot of work in machine learning does not consider the form above, instead, they simplify this kind of problem into a binary classification problem. They further explain the problems. First, a ranking function could simply get a really high prediction accuracy by just classifying all data into irrelevant ones as there are always a lot more irrelevant documents. At the same time, sometimes we don't have an absolute standard of relevance for data. Under many circumstances, users just want to get the 
most relevant document instead of all relevant documents. Therefore, he proposes an empirical risk minimization approach in his work. Given an independently and identically distributed training sample $S$ of size $n$ containing queries $q$ with their target rankings $r^*$
\begin{equation}
    (q_1,r_1^*),(q_2,r_2^*),\dots,(q_n,r_n^*)
\end{equation}
And the Learner $L$ will select a ranking function f from a family of ranking functions $F$ that maximizes the empirical $\tau$
\begin{equation}
    \tau_S(f)=\frac{1}{n}\sum^n_{i=1}\tau(r_{f(q_i),r_i^*)}
\end{equation}
on the training samples. And this optimization goal can be realized by minimizing the number of discordant pairs, and can be approximated by introducing slack variales.

The threshold learning and regularization in RankSVMs have also been well studied. It is also
pointed out in some paper \cite{WU202024} that combining binary relevance and Rank SVM has a satisfactory effect as RankSVM could minimize Hamming distance while the binary relevance can remove the need for extra threshold learning. So in our work, we are inspired to use RankSVM with binary relevance. 

\subsection{Language Models}
As we have sketched in the introduction, there has been a lot of work in improving language model architecture. Many researchers have provided their version of BERT\cite{devlin2019bertpretrainingdeepbidirectional} variants incorporating legal elements. Lawformer\cite{xiao2021lawformer} is a meaningful endeavor in pre-trained model breakthroughs. Also, other than the common practice of using transformer-based models, a new model architecture MAMBA\cite{gu2024mambalineartimesequencemodeling} has also shown attractive performance, especially for long-input NLP tasks. So in our work, we plan to research four kinds of models: the original BERT model, the BERT model used in the LEVEN baseline dataset, the original Lawformer model, and MAMBA. We wish to use these four kinds of representative models to tell their difference in feature extraction in the SCR task and reach a conclusion about how model architecture may affect the ranking.

\section{Methodology}

\subsection{Overview}
In this part, we are to give an introduction to the similar case retrieval task and the steps we take in our model. Similar to most document retrieval tasks, similar case retrieval aims to get candidate cases similar to query cases from a candidate case pool, and they are ranked according to the similarity. Formally, given a query case $q$ and a candidate case set $C=\{c_i|i=1,\dots, n\}$, the rank we get for a candidate file $c_i\in C$ is $R_{c_1/q}$; then the most relevant candidate case is 
\begin{equation}
    c_{*/q}=arg\ max\{R_{c_i/q}| c_i \in C\}
\end{equation}

Though there are a lot of parts in a judgment document, we only consider the fact part in a document. Our inputs consist of one query and one candidate. We use language models to extract features from these inputs and send them into our ranking functions.
\subsection{Data Preprocessing}
\begin{figure}
  \centering
  \includegraphics[width=\linewidth]{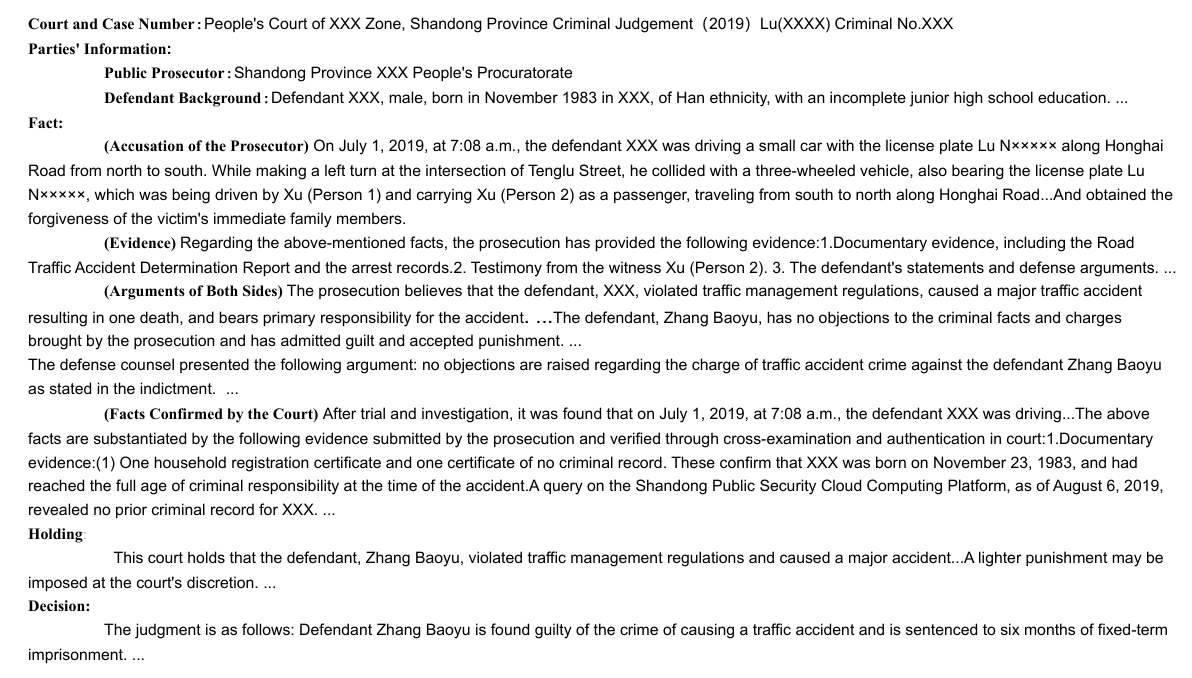}
  \caption{Architecture of a legal judgement}
  \label{case}
\end{figure}

In our experiments, the documents we use all come from China Judgements Online; they can usually be partitioned into 3 parts: the information of two parties involved, the facts part, and the decisions made by the court, as is shown in Figure \ref{case}. Here we
follow the common practice to only take fact parts of those documents as inputs, see
equation \ref{input} and \ref{input_v2} respectively for LeCaRDv1 and LeCaRDv2.
\begin{equation}
    input_{v1}=(query['q'],cand['ajjbqk'])
    \label{input}
\end{equation}
\begin{equation}
    input_{v2}=(query['fact'],cand['fact'])
    \label{input_v2}
\end{equation}

\subsection{Fine-tuning with Pretrained Model and Feature Extraction}
We need language models to capture the deep semantic representation in lagel texts so we conduct fine-tuning first. 
When we do fine-tuning, we basically follow the framework taken in LEVEN code
\url{https://github.com/thunlp/LEVEN}. We add a two-class classification layer after the
pooled outputs denoted as [CLS] or the hidden state and do fine-tuning with the training set, see equation \ref{feature}. The
hyperparameters used are shown in Table \ref{table:Parameters for BERT model}.
\begin{equation}
output=func_{twoclassclassification}([CLS])
\label{feature}
\end{equation}

As we said before, we decided to use binary relevance labels instead of considering different extents of relevance in our model. According to the description of LeCaRDv1 and LeCaRDv2 papers, relevance could be classified into 4 classes. However, in our experiments, we only consider golden labels( relevance greater than 2) to be positive labels when training, following common practice. 

To extract features, we just use the output of CLS of every query-candidate pair
as our features.  For further experiments, we only pick the models with the best
performance on the validation dataset or test dataset for feature extraction. While
after training and validating on each fold, we report the average value for the 5 folds
on LeCaRDv1, we report the performance for the test dataset on LeCaRDv2. In
both datasets,when training on each fold,we iterate the training dataset for 5 epochs
and choose the epoch with the best performance compared with other epochs on the
respective validation/test dataset.

The features extracted are used to send into ranking functions,stated as follows, see equation \ref{feature}.
\begin{equation}
 features=[CLS]
 \label{feature}
\end{equation}
\subsection{RankSVM}

Different forms of RankSVM are suitable for different types of data. In our experiments, we follow the 1-slack algorithm that is solved with the dual problem, which according to our experiments, is better than other settings\cite{Joachims1998MakingLS}. The basic form of RankSVM is shown as follows.
\begin{equation}min\frac{1}{2}\||w\||^2+C\sum^n_{i=1}\sum_{u,v:y^{(i)}_{u,v}=1}\xi_{u,v}^{(i)}\end{equation}
\begin{equation}s.t. w^T(x_u^{(i)}-x_v^{(i)})\geq1-\xi_{u,v}^{(i)}, if y_{u,v}^{(i)}=1\end{equation},
\begin{equation}\xi_{u,v}^{(i)}\geq 0,i=1,\dots,n.
\end{equation}

To improve the efficiency, this method also endeavors to choose the direction that could help this iteration find the solution faster. It has also tried to remove those examples that apparently do not affect the support vectors when training. 

The efficiency in this algorithm further exists in its cutting-plane solving methods. As all constraints share one slack variant, its bound can easily be verified using its dual problem, written as follows:
\begin{equation}
     \max_{\alpha \geq 0}  \sum_{c \in [0,1]^n} \frac{\|c\|_1}{n} \alpha_c - \frac{1}{2} \sum_{c \in [0,1]^n} \sum_{c' \in [0,1]^n} \alpha_c \alpha_{c'} x_c^T x_{c'}
\end{equation}
\begin{equation}
    \text{s.t.}\sum_{c \in [0,1]^n} \alpha_c \leq C
\end{equation}

It could swiftly get the $\epsilon$-approximate solution, and could be solved within constant times of iteration. The time it takes in classification problem is only $O(sn)$. It is also very efficient for large dataset\cite{10.1145/1150402.1150429}.

 To tune how soft the
classifier is, we use different values of C, and they are 0.001, 0.05, 0.01, 0.02, 0.05, 0.1,
0.5, 1, 10, 100.  In our experiments, we only use the linear kernel. Here we only report
the best result for each model to compare with the traditional fully-connected layer.  

\section{Experiments}
\subsection{Experiments Settings}
we use two benchmark datasets that are commonly used for similar
case retrieval tasks–LeCaRDv1 and LeCaRDv2. Here we mainly examine the difference
between these two datasets in data distribution. LeCaRDv1 dataset(A Chinese Legal Case Retrieval
Dataset)\cite{ma2021lecard} and LeCaRDv2 dataset\cite{li2023lecardv2largescalechineselegal} are two datasets
specifically and widely used as benchmarks for similar case retrieval tasks since their
release. The details of these two are shown in Table \ref{tab:details}. LeCaRDv1 contains
107 queries and 10,700 candidates, among which each query only can retrieve 100
candidates that belong to the respective subfolder. LeCaRDv2 has 800 queries and each
query can retrieve from the all-in-one corpus containing 55,192 candidates. At the same time, the ratio of relevant cases is much lower in LeCaRDv2. 

\begin{figure}[h]
     \centering
     \begin{subfigure}[b]{0.45\textwidth}
         \centering
         \includegraphics[width=\textwidth]{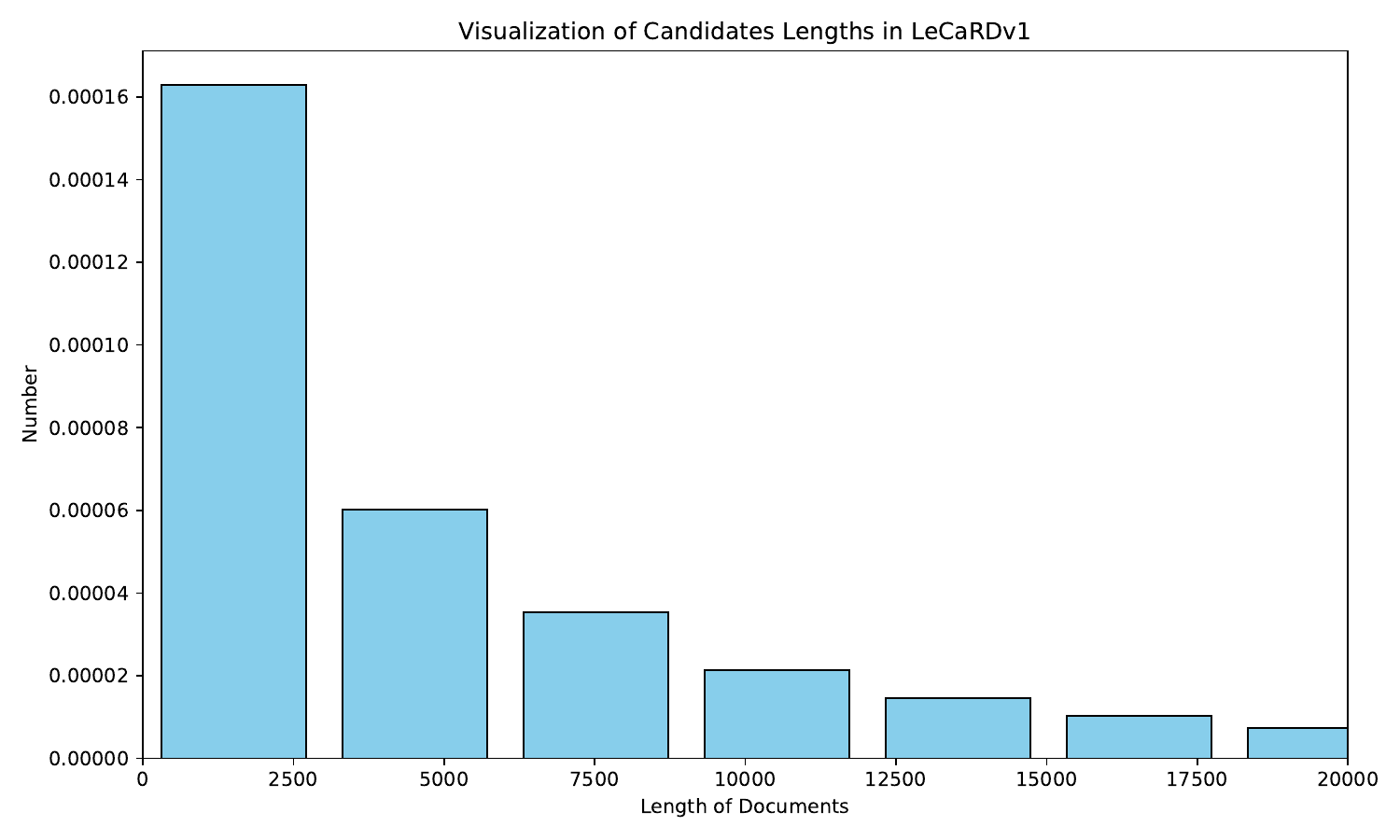}
         \caption{Lengths of Candidates in LeCaRDv1 Dataset}
         \label{fig:Lengths of Candidates in LeCaRDv1}
     \end{subfigure}
     \hfill
     \begin{subfigure}[b]{0.45\textwidth}
         \centering
         \includegraphics[width=\textwidth]{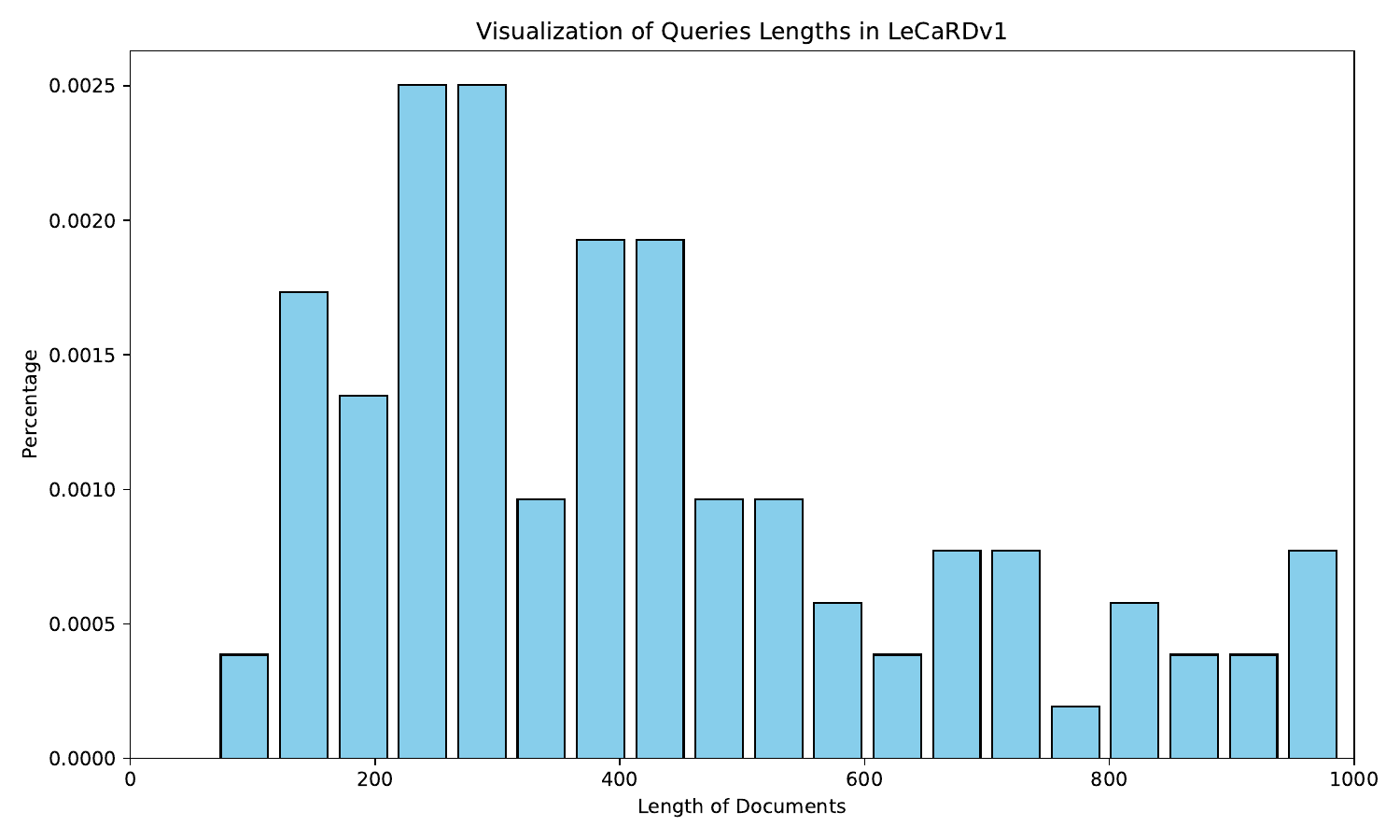}
         \caption{Lengths of Queries in LeCaRDv1 Dataset}
         \label{fig:Lengths of Queries in LeCaRDv1 Dataset}
     \end{subfigure}
        \caption{Length of Documents in LeCaRDv1 Dataset}
        \label{fig:lengths of datasets}
\end{figure}

\begin{figure}
     \centering
     \begin{subfigure}[b]{0.45\textwidth}
         \centering
         \includegraphics[width=\textwidth]{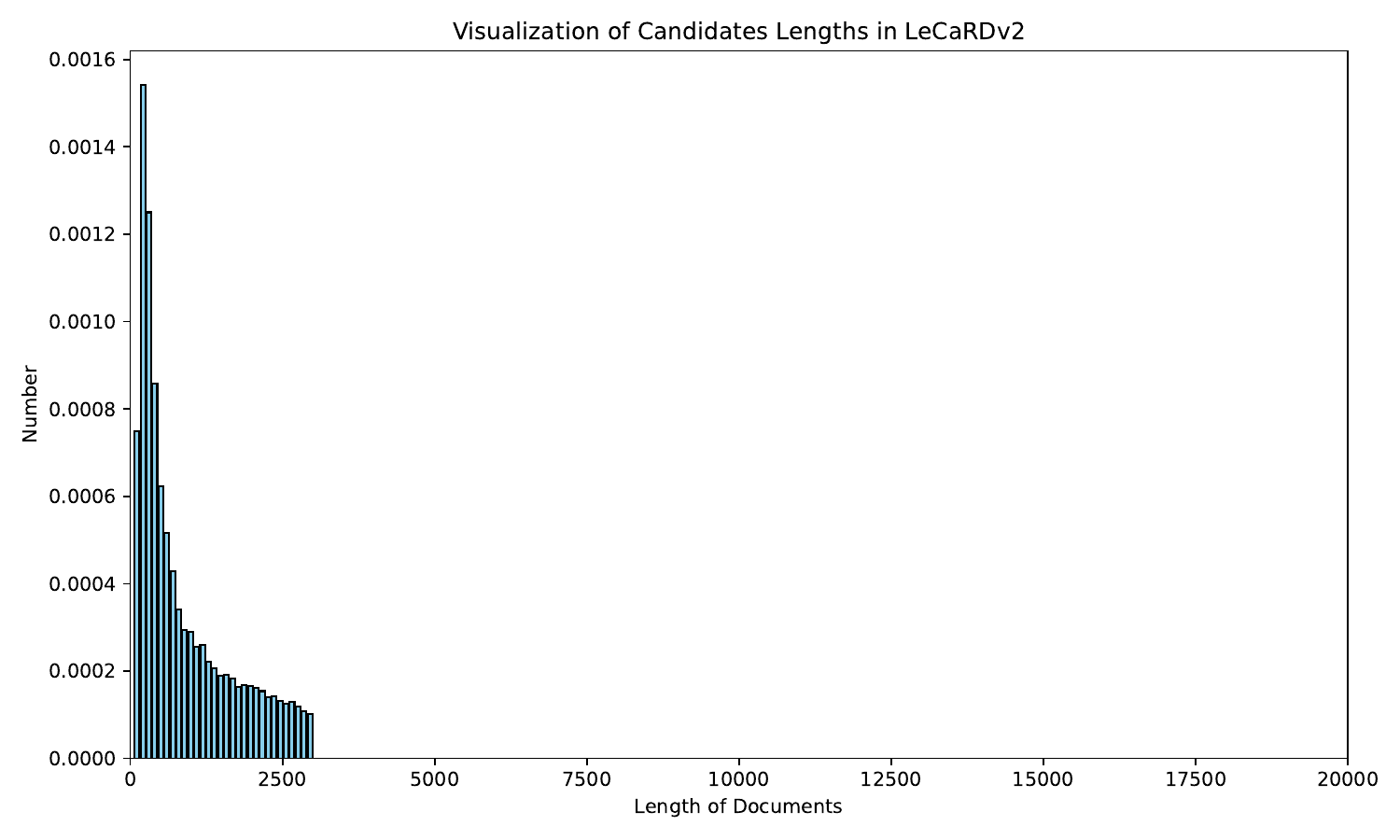}
         \caption{Lengths of Candidates in LeCaRDv2 Dataset}
         \label{fig:Lengths of Candidates in LeCaRDv2}
     \end{subfigure}
     \hfill
     \begin{subfigure}[b]{0.45\textwidth}
         \centering
         \includegraphics[width=\textwidth]{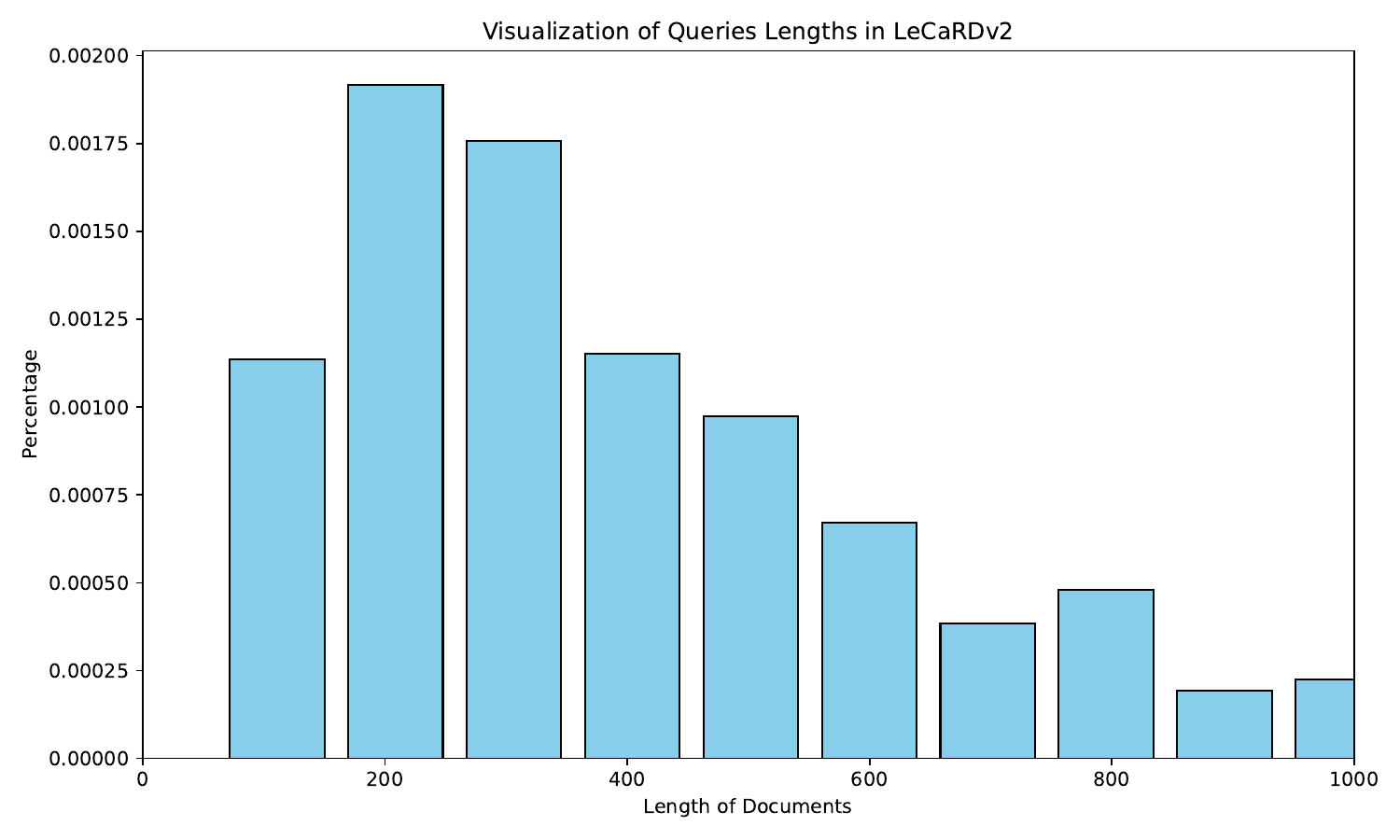}
         \caption{Lengths of Queries in LeCaRDv2 Dataset}
         \label{fig:Lengths of Queries in LeCaRDv2 Dataset}
     \end{subfigure}
        \caption{Length of Documents in LeCaRDv2 Dataset}
        \label{fig:lengths of datasets_v2}
\end{figure}

At the same time, it is worth attention that while LeCaRDv1 has a separate candidate
pool(folder) of size 100 for each query, there is no subfolder for LeCaRDv2 queries.
Moreover, there lies some divergence in their judging criteria. While LeCaRDv1
focuses on the fact part of a case involving key circumstances and key elements,
LeCaRDv2 argues that characterization, penalty, and procedure should all be taken
into consideration, even though they say in their paper that there is no explicit
mapping function between the Overall Relevance and the sub-relevance.
\begin{table*}
\caption{Details about LeCaRDv1 and LeCaRDv2 datasets}
\centering
\begin{tabular}{c c c} 
\toprule
datasets&\textbf{LeCaRDv1}	& \textbf{LeCaRDv2}\\
\midrule
\#candidate cases/query		& 100&55192\\
\midrule
\#average relevant cases per query&10.33&20.89\\	
\midrule
ratio of relevance candidates(positive)&0.1033&0.0004\\
\bottomrule
\end{tabular}
\label{tab:details}
\end{table*}
Moreover, we examine the fact parts in both datasets to be used as our inputs,
see Figure \ref{fig:lengths of datasets} and Figure \ref{fig:lengths of datasets_v2}, from which we could tell that many of them are actually
longer than the input limit of the BERT model as well as the Lawformer or MAMBA
model. At the same time, as we use the same number of bins in the histogram for
both LeCaRDv1 and LeCaRDv2 datasets, it could be seen that while the LeCaRDv2
dataset has query files of longer lengths, it has candidate files of shorter lengths, which may be due to the way the dataset is built.

With different benchmark datasets, we conduct different data splitting. As there
is no subfolder for the LeCaRDv2 dataset in its original repository when our experiments are conducted \footnote{https://github.
com/THUIR/LeCaRDv2} and there are only labels for 30 candidate cases concerning a query, we can hardly train our model according to our experiments just using these data. On the other hand, out of consideration for memory and the potential risk of probable overfitting if we conduct experiments with all negative samples among the
corpus, we conduct our experiments trying to follow the practice on the LeCaRDv1
dataset by building a subfolder of 130 candidate files for each query, among which 30
are labeled relevant, while the rest are sampled from the large candidate corpus. This is also similar to the practice taken in the
LeCaRDv2 baseline experiment where  the ratio of positive samples and
negative samples is 1:32. It is shown by our result that it is a wise choice. We also follow
different train-test splits. For LeCaRDv1, we follow the 5-fold validation, while for
LeCaRDv2 we took 640 instances as our training dataset and 160 as our test dataset.

At the same time, the input length does differ among models. For
BERT-based models, we truncate the candidate input lengths to 409 and the
query input lengths to 100, and the format could be checked in Equation \ref{BERT} with special tokens. For
Lawformer, we basically follow the practice in pretraining to truncate lengths to respectively 3072 and 509, and the format is the same as BERT. For MAMBA,
the length of candidate files is 500 and the length of query files is 300,see Equation \ref{Mamba}.
\begin{equation}
    inputx_{BERT}=[CLS]{query}[SEP]{cand}[SEP]
    \label{BERT}
\end{equation}
\begin{equation}
inputx_{Mamba}=document:\{cand\}\backslash n \backslash n query:\{query\}[EOS]
\label{Mamba}
\end{equation}

Regarding hyperparameters, we don't spend a lot of time tuning them and basically follow LEVEN's code; they could be checked in \ref{table:Parameters for BERT model}.
\begin{table*}
\caption{Parameters of Different Models}
\centering
\begin{tabular}{ c c c c}
\toprule
Model&BERT-Based&Lawformer&MAMBA\\
\midrule
  Algorithm& 1-slack algorithm (dual)& 1-slack algorithm (dual)& 1-slack algorithm (dual) \\
\midrule
Norm&l1-norm&l1-norm&l1-norm\\
\midrule
Kernel&Linear&Linear&Linear\\
\midrule
Learning Rate&1e-5&2.5e-6(-1)/5e-6(v2)&1e-5\\
\midrule
Optimizer&Adamw&Adamw&Adamw\\
\midrule
Training batch size&16&2(v1)/4(v2)&16\\
\midrule
Evaluating batch size&32&32&32\\
\midrule
Step size&1&1&1\\
\bottomrule
\end{tabular}
\label{table:Parameters for BERT model}
\end{table*}
\subsection{Experiments on LeCaRDv1}
To test the effectiveness of our method, we conduct our experiments mainly using
BERT, BERT+Event, BERT+Event+RankSVM, BERT+RankSVM, Lawformer,
Lawformer+RankSVM,Mamba and Mamba+RankSVM. In our experiments, we take NDCG as our main evaluation metric, which is a common metric in SCR task. It considers the relevance of the result. The result is shown in Table
\ref{tab:tab3}.
\begin{table*}
  \caption{Results on LeCaRDv1 Dataset}
  \label{tab:tab3}
  \begin{tabular}{ccccc}
    \toprule
    Model&NDCG@10&NDCG@20&NDCG@30&P@5\\
    \midrule
    BERT & 0.7896& 0.8389&0.9113&0.4556\\
    BERT+RankSVM & \textbf{0.7963}& \textbf{0.8504}&\textbf{0.9166}&\textbf{0.4571}\\
    \midrule
    BERT+LEVEN& 0.7881 & 0.8435&0.9136&0.4637\\
     BERT+LEVEN+RankSVM&\textbf{0.7931}& \textbf{0.8452}&\textbf{0.9164}&\textbf{0.4717}\\
     \midrule
     MAMBA &0.7136&\textbf{0.7760}&\textbf{0.8744}&0.3888\\
     MAMBA+RankSVM&\textbf{0.7144}&0.7743&0.8742&\textbf{0.4002}\\
     \midrule
     Lawformer&0.7592&0.8169&0.8993&\textbf{0.4081}\\
     Lawformer+RankSVM&\textbf{0.7738}&\textbf{0.8258}&\textbf{0.9073}&0.4033\\
  \bottomrule
\end{tabular}
\end{table*}
From the results, we could see that RankSVM helps improve performance on
NDCG and has exceptional performance on precision for the top 5. And the advantage in NDCG is more obvious for the more relevant cases, which proves that our model is better
at putting highly relevant cases earlier in a similar case list. To better understand
the advantage of RankSVM, we finish visualization based on the training and testing
dataset on fold 0 to see the score computed by original BERT and BERT+RankSVM.

\begin{figure}[ht]
     \centering
     \begin{subfigure}[b]{0.32\textwidth}
         \centering
         \includegraphics[width=\textwidth]{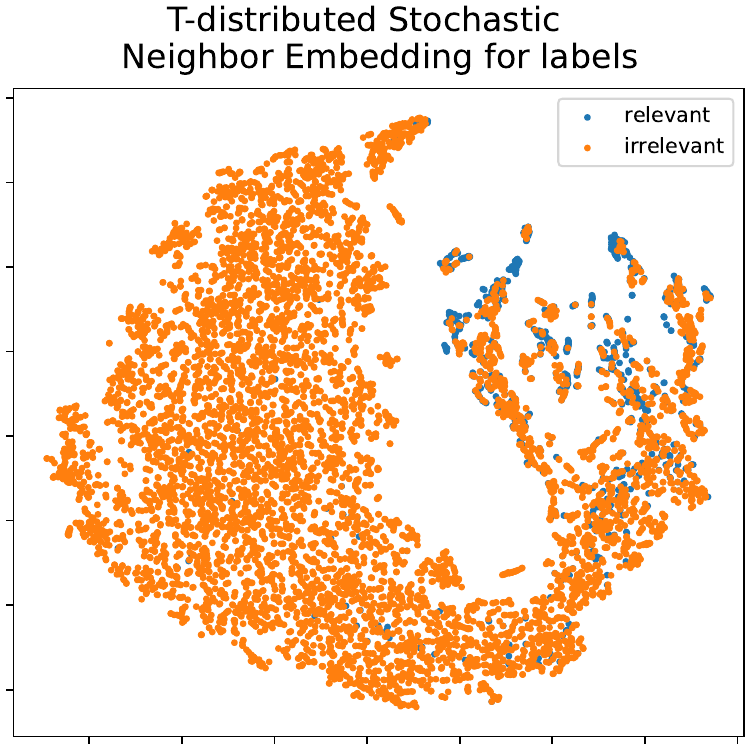}
         \caption{Visualization of Relevance Labels}
         \label{fig:labels_v1}
     \end{subfigure}
     \hfill
     \begin{subfigure}[b]{0.32\textwidth}
         \centering
         \includegraphics[width=\textwidth]{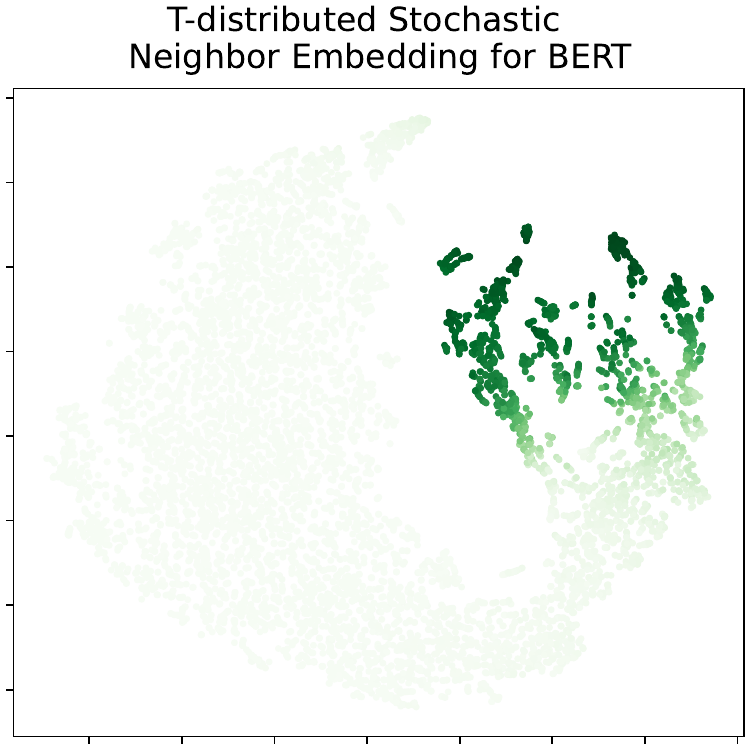}
         \caption{Visualization of BERT Scores}
         \label{fig:BERT_v1}
     \end{subfigure}
     \hfill
     \begin{subfigure}[b]{0.32\textwidth}
         \centering
         \includegraphics[width=\textwidth]{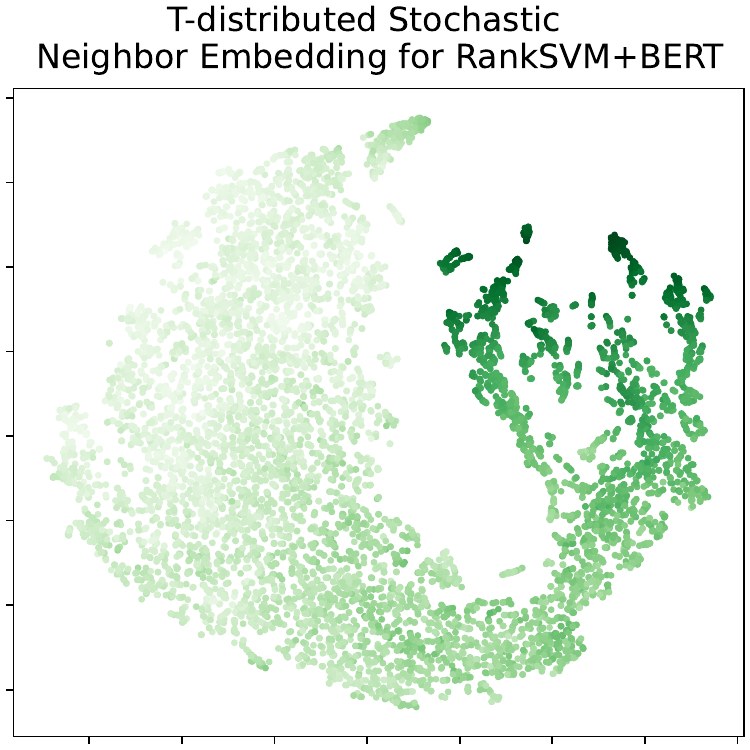}
         \caption{Visualization of RankSVM Scores}
         \label{fig:RankSVM_v1}
     \end{subfigure}
        \caption{Visualization of Labels/Features/Scores on LeCaRDv1 Training Dataset}
        \label{fig:v1_train}
\end{figure}

\begin{figure}
     \centering
     \begin{subfigure}[b]{0.32\textwidth}
         \centering
         \includegraphics[width=\textwidth]{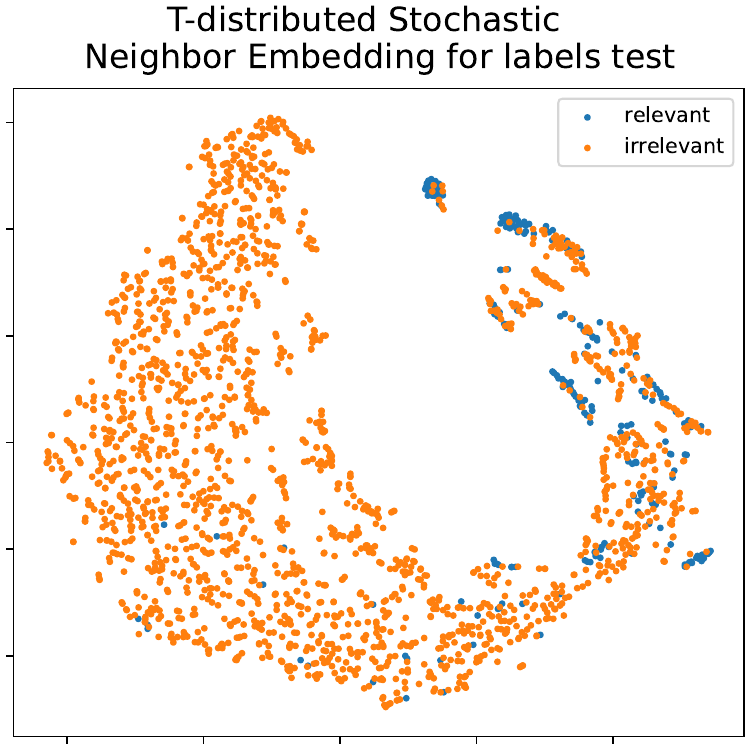}
         \caption{Visualization of Relevance Labels}
         \label{fig:labels_v1_test}
     \end{subfigure}
     \hfill
     \begin{subfigure}[b]{0.32\textwidth}
         \centering
         \includegraphics[width=\textwidth]{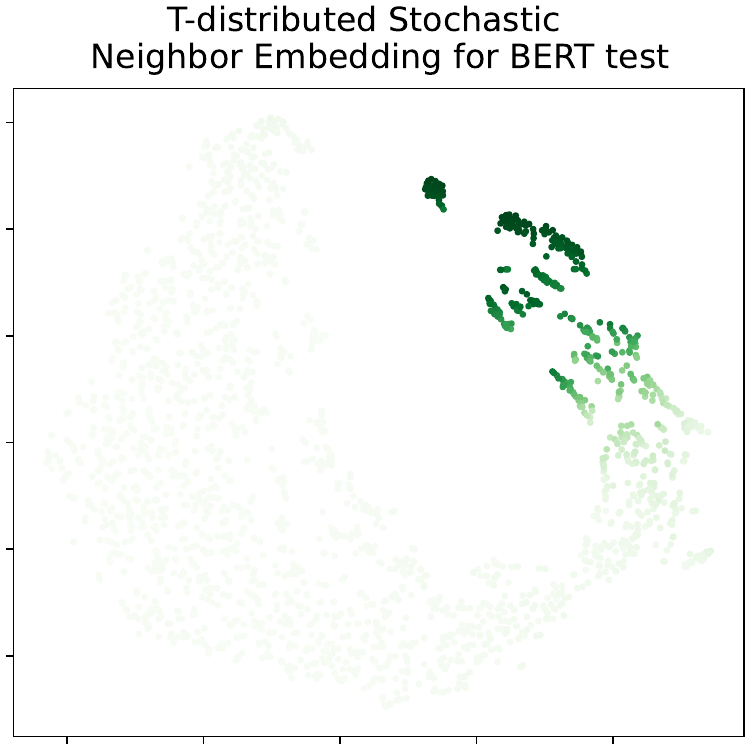}
         \caption{Visualization of BERT Features}
         \label{fig:BERT_test_v1}
     \end{subfigure}
     \hfill
     \begin{subfigure}[b]{0.32\textwidth}
         \centering
         \includegraphics[width=\textwidth]{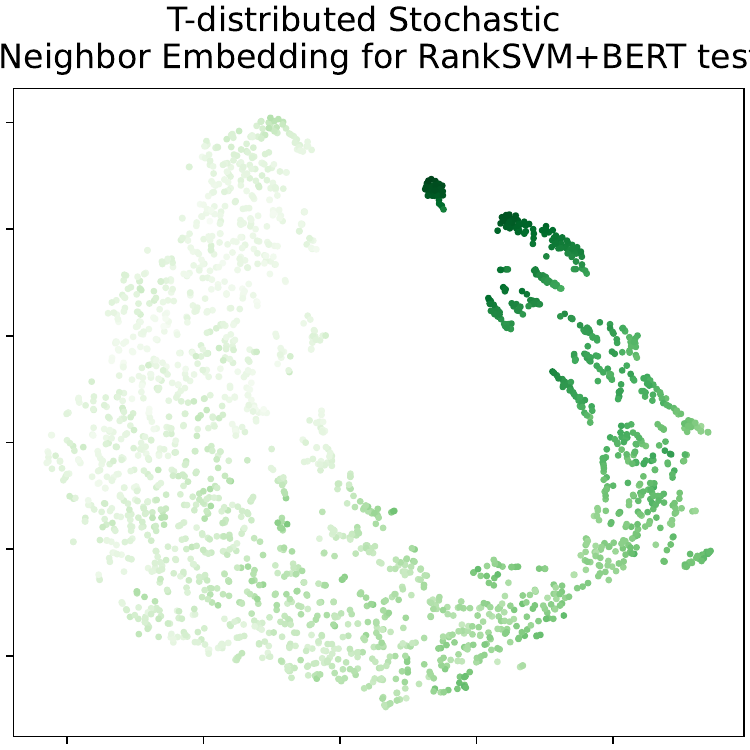}
         \caption{Visualization of RankSVM}
         \label{fig:RankSVM_v1_test}
     \end{subfigure}
        \caption{Visualization of Labels/Features/Scores on LeCaRDv1 Test Dataset}
        \label{fig:v1_test}
\end{figure}

Specifically, we use features extracted by BERT as our input and compress them into
two dimensions using t-SNE\cite{JMLR:v9:vandermaaten08a}. After that, we color
them according to similarity scores computed by relative models or labels. See Figure
\ref{fig:v1_train} and \ref{fig:v1_test} for reference.

From the visualization result, we could see that RankSVM performs better by
differentiating the relevance level of each pair continuously, while BERT treats the
relevance level more discretely.

\begin{figure}
     \centering
     \begin{subfigure}[b]{0.47\textwidth}
         \centering
         \includegraphics[width=\textwidth]{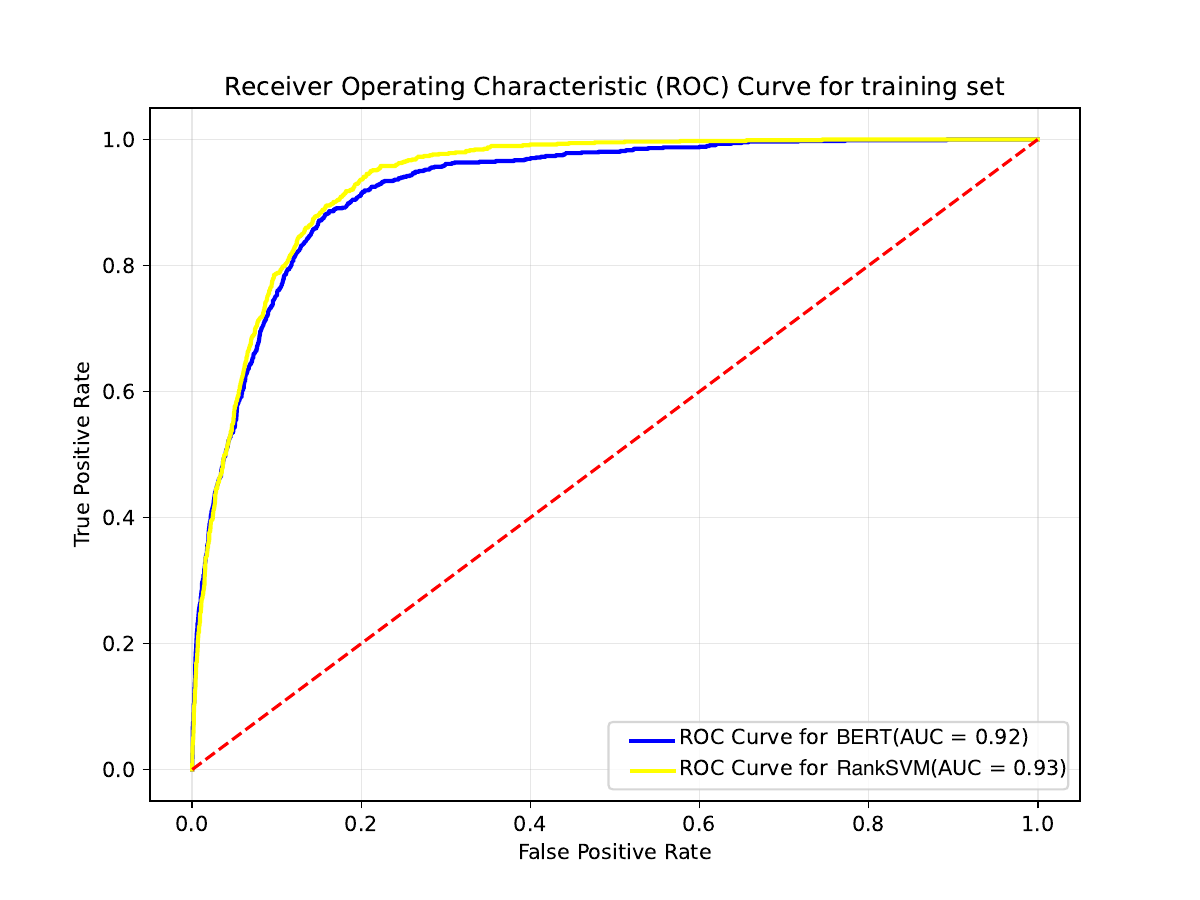}
         \caption{ROC Curve for LeCaRDv1 Training Dataset with BERT based models}
         \label{fig:ROC_v1_train_BERT}
     \end{subfigure}
     \hfill
     \begin{subfigure}[b]{0.47\textwidth}
         \centering
         \includegraphics[width=\textwidth]{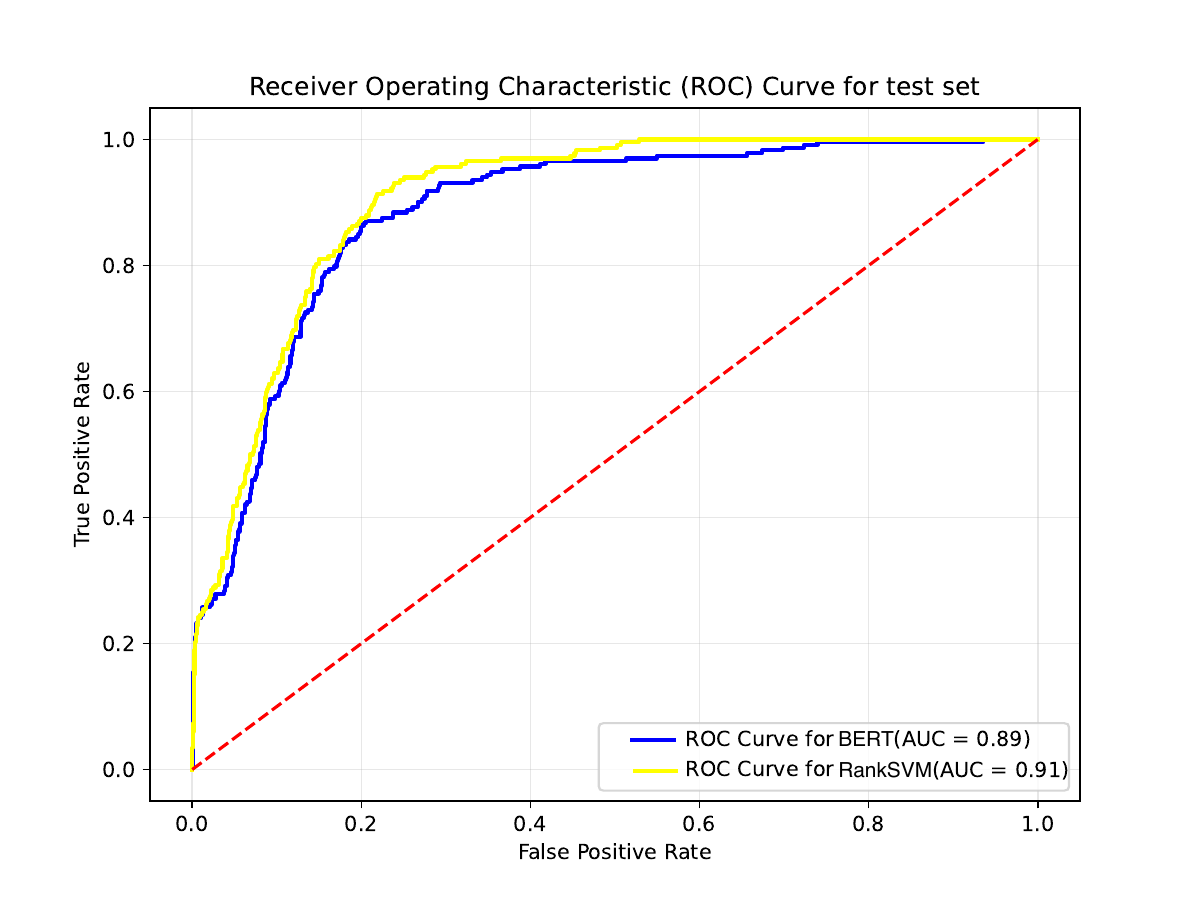}
         \caption{ROC Curve for LeCaRDv1 Test\\ Dataset with BERT based models}
         \label{fig:ROC_v1_test_BERT}
     \end{subfigure}
        \caption{ROC Curves Using BERT model and BERT+RankSVM}
        \label{fig:ROC_curve_v1}
\end{figure}

\begin{figure}
     \centering
     \begin{subfigure}[b]{0.47\textwidth}
         \centering
         \includegraphics[width=\textwidth]{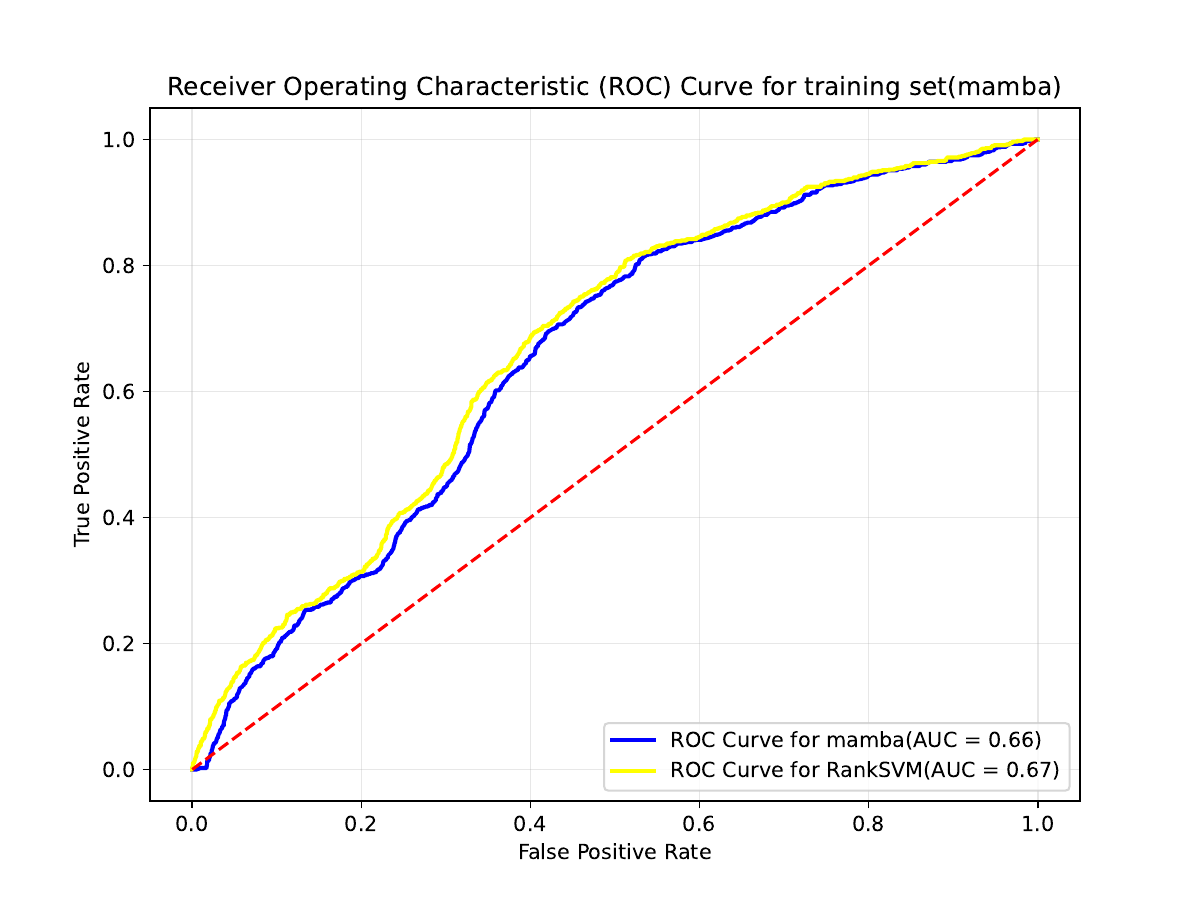}
         \caption{ROC Curve for LeCaRDv1 Training Dataset with MAMBA based models}
         \label{fig:ROC_v1_train}
     \end{subfigure}
     \hfill
     \begin{subfigure}[b]{0.47\textwidth}
         \centering
         \includegraphics[width=\textwidth]{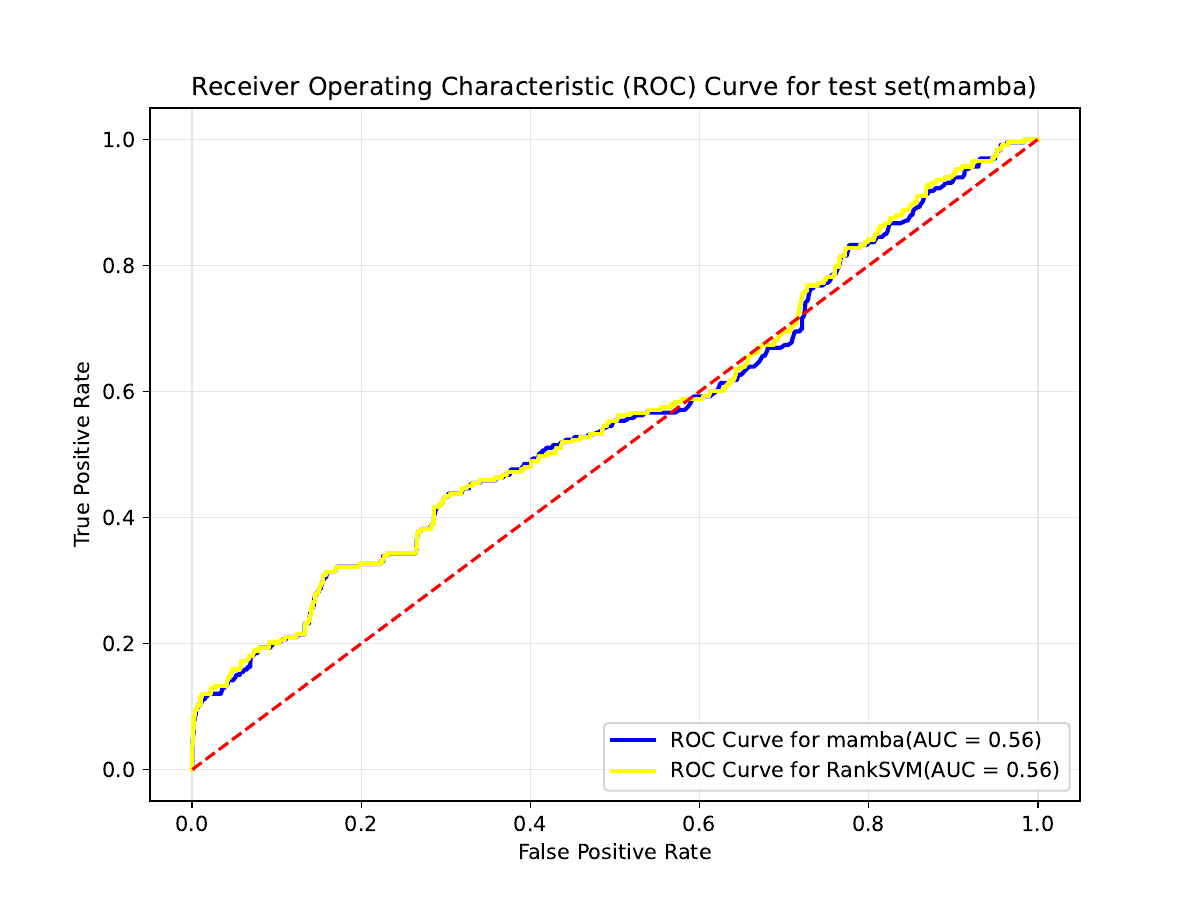}
         \caption{ROC Curve for LeCaRDv1 Test\\ Dataset with MAMBA based models}
         \label{fig:ROC_v1_test}
     \end{subfigure}
        \caption{ROC Curves Using MAMBA model and MAMBA+RankSVM}
        \label{fig:ROC_curve_v1 on mamba}
\end{figure}

What’s more, the mechanism of RankSVM is actually maximizing ROC(Receiver
Operating Characteristic curve) pointed out in some early paper \cite{Ataman2005OptimizingAU}, said to outperform linear SVMs, perceptron and logit. The AUC is especially suitable for class imbalance setting. So we use BERT with RankSVM and BERT without RankSVM to compare their ROC and AUC values. We use the first fold data and the C is 0.01. The visualization result
is shown in Figure \ref{fig:ROC_curve_v1} and \ref{fig:ROC_curve_v1 on mamba}. It could be seen that RankSVM helps improve
the result of the BERT model on the AUC score, and it does even better in the test
dataset. It proves our guess before that it could help reduce overfitting without sacrificing efficiency.

\subsection{Experiments on LeCaRDv2}
The result on LeCaRDv2 is shown in Table \ref{table:LeCaRDv2 Result}. As the performance of MAMBA is not as competitive in LeCaRDv1, we don't use it for LeCaRDv2. According to the table, we could find that the NDCG
related metrics are generally improved similar to LeCaRDv1, except for the BERT
model. 

\begin{figure}
     \centering
     \begin{subfigure}[b]{0.45\textwidth}
         \centering
         \includegraphics[width=\textwidth]{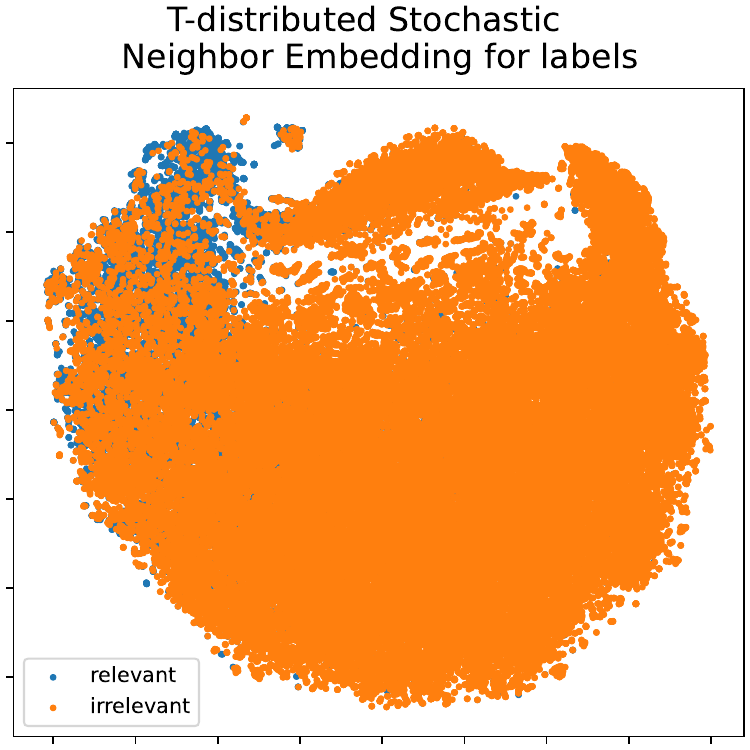}
         \caption{Visualization of Relevance Labels}
         \label{fig:v2_label}
     \end{subfigure}
     \hfill
     \begin{subfigure}[b]{0.45\textwidth}
         \centering
         \includegraphics[width=\textwidth]{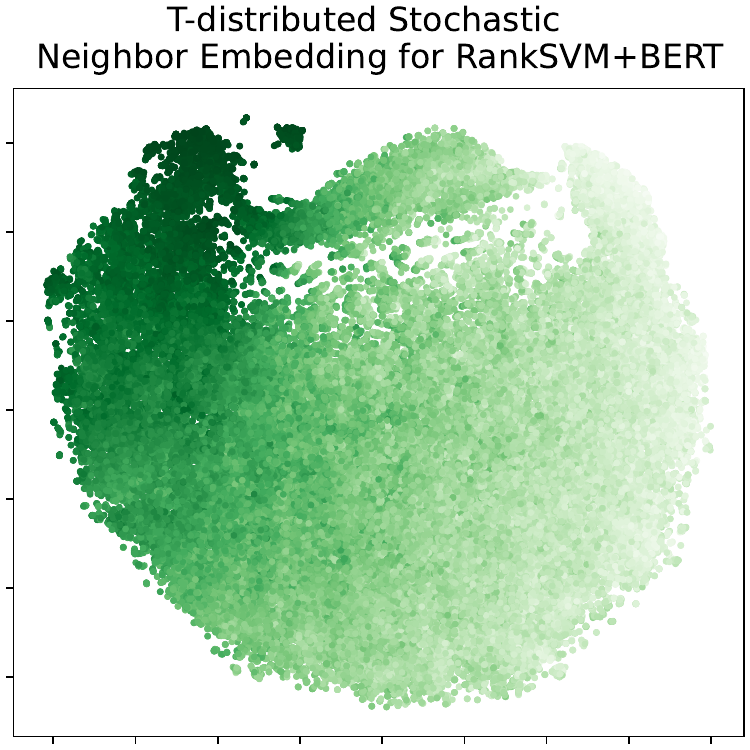}
         \caption{Visualization of RankSVM Scores}
         \label{fig:v2_SVM}
     \end{subfigure}
        \caption{Visualization of Labels/Scores on LeCaRDv2 Training Dataset}
        \label{fig:Visualization_on_v2_train}
\end{figure}

\begin{figure}
     \centering
     \begin{subfigure}[b]{0.45\textwidth}
         \centering
         \includegraphics[width=\textwidth]{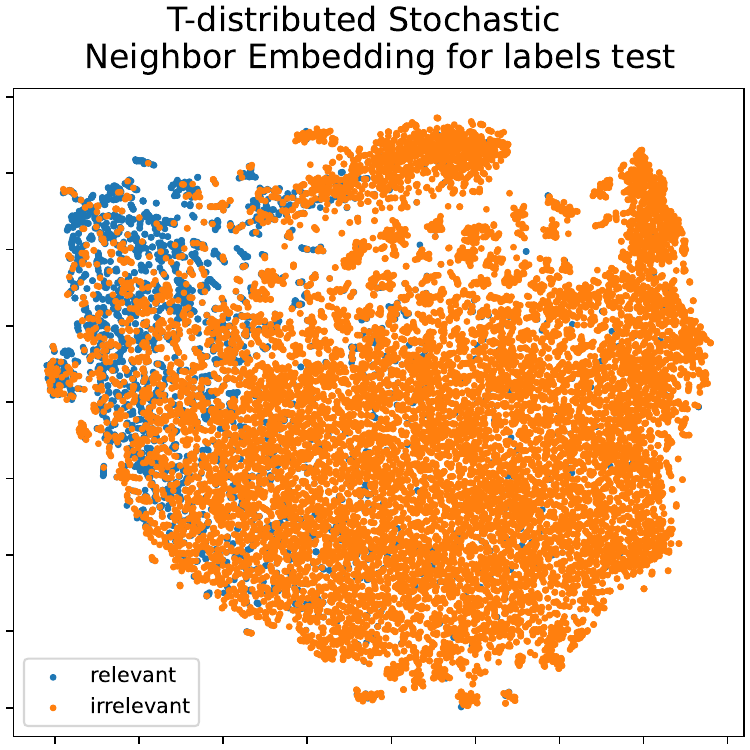}
         \caption{Visualization of Relevance Labels}
         \label{fig:v2_label_test}
     \end{subfigure}
     \hfill
     \begin{subfigure}[b]{0.45\textwidth}
         \centering
         \includegraphics[width=\textwidth]{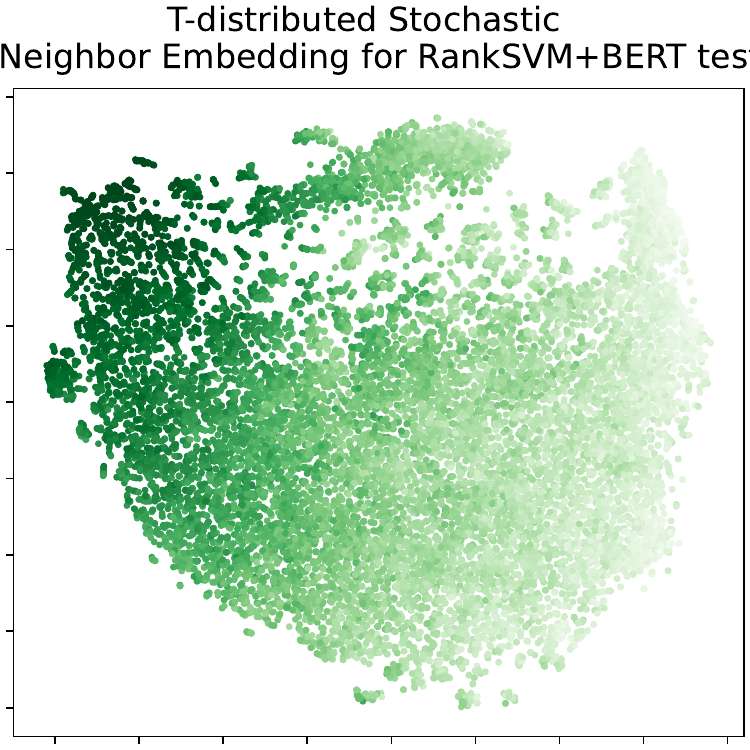}
         \caption{Visualization of RankSVM Scores}
         \label{fig:v2_SVM_test}
     \end{subfigure}
        \caption{Visualization of Labels/Score on LeCaRDv2 Test Dataset}
        \label{fig:Visualization_on_v2_test}
\end{figure}

To understand its performance, here we also do visualization using features
extracted from the BERT model following the same steps as the LeCaRDv1 dataset,
see Figure \ref{fig:Visualization_on_v2_train} and \ref{fig:Visualization_on_v2_test}.
\begin{table*}
\caption{Results on LeCaRDv2 Dataset}
\centering

\begin{tabular}{c c c c c} 
\toprule
 Model&NDCG@10&NDCG@20&NDCG@30\\
\midrule
 BERT&\textbf{0.8351}&\textbf{0.8764}&\textbf{0.9348}\\
 BERT+RankSVM&0.793&0.8582&0.9247\\
\midrule
BERT+LEVEN&\textbf{0.8117}&0.8455&0.9184 \\
BERT+LEVEN+RankSVM&0.8078&\textbf{0.8548}&\textbf{0.9227}\\
\midrule
Lawformer&0.7968&0.8461&0.919\\
Lawformer+RankSVM&\textbf{0.812}&\textbf{0.8574}&\textbf{0.9247}\\
\bottomrule
\end{tabular}
\label{table:LeCaRDv2 Result}
\end{table*}

It is shown by the figure with real labels that BERT features fail to distinguish
these two classes explicitly this time as there are a lot of overlaps between the two classes compared with what is extracted from the LeCaRDv1 dataset, while linear RankSVM
tries to learn from those features that are not distinguished between these two classes
continuously. The failure of BERT may be owing to the changed standards of relevance
in LeCaRDv2 and the shorter document length of the LeCaRDv2 dataset. The visualization explains
why RankSVM added to BERT fails to help improve NDCG metrics here.
\subsection{Comparison Study}
To further explore the features of RankSVM and the potential of pairwise ranking
methods.We finish comparison experiments with RankNet using some fraction of data
from LeCaRDv1, and the result is shown as follows, see table \ref{tab:compare}. We can tell that RankNet
is also very competitive especially considering its precision and it proves that there is a lot to explore with pairwise
methods.

\begin{table*}
    \centering
    \begin{tabular}{c c c c c c }
        \hline
         Methods & NDCG@10 & NDCG@20 & NDCG@30 & P5 & P10 \\ \hline
        RankSVM & 0.6995 & 0.7711 & 0.8576 & 0.3 & 0.29 \\ \hline
        RankNet & 0.7816 & 0.8336 &  0.8432 & 0.5 & 0.62 \\ \hline
        
    \end{tabular}
    \caption{Results on Comparison Experiments with RankNet}
    \label{tab:compare}
\end{table*}
\section{Conclusion}

Even after we changed the length of the input data from 512 to 800 using the MAMBA
model and Lawformer model, we could see that there
is no apparent increase in metrics compared with BERT, which is also mentioned in
other works\cite{deng2024learning}. It is worth exploring how long-input language models could really benefit downstream applications.

To investigate the reason why RankSVM fails to improve the performance on BERT-based
models on the LeCaRDv2 dataset, we conduct experiments using multi-class
BERT classification to extract features.  We speculate that RankSVM on
LeCaRDv2 needs language models that provide more information for ranking as Lawformer and BERT+LEVEN model have provided more information. We end up getting
all NDCG-related metrics improved again, see table \ref{table:Result when using multi-class classification}. RankSVM still works for BERT on the LeCaRDv2 dataset under this setting.

\begin{table*}
\caption{Result on LeCaRDv2 when we do multi-class classification}
\centering
\begin{tabular}{c c c c}
\toprule
Model&NDCG@10&NDCG@20&NDCG@30\\
\midrule
  BERT Multiclass&0.791&0.8443&0.9181 \\
\midrule
BERT Multiclass+RankSVM&0.8053&0.8352&0.9222\\
\bottomrule
\end{tabular}
\label{table:Result when using multi-class classification}
\end{table*}

In our paper, we examine the performance of RankSVM when combined with other
language models using binary labels on SCR tasks on the
LeCardv1 dataset and the LeCardv2 dataset. We come to the conclusion that
it could help improve the performance of models by
improving their concrete ranks, and we also give explanations from the
feature extraction perspective. The improvements are more obvious for more relevant documents. Also, the RankSVM could help mitigate overfitting with surprising efficiency, which results
from an imbalance class common in this task. Our findings point to the potential of
improving SCR task performance from a learning-to-rank perspective, especially considering pairwise methods. 

While previous work has considered different ranking functions, there has been no research focus on their difference and their suitability for SCR tasks, especially pairwise methods. There still lies  
some work that needs to be done to explore the appropriate ranking function to improve performance
on the recall metric on the LeCaRDv2 dataset and how to exploit long inputs to improve performance. We also leave
the investigation regarding other pairwise methods combined with other models and datasets for
future work.


\section*{Declaration on Generative AI}
 During the preparation of this work, Yuqi Liu used the GPT model in order to: Grammar and spelling check. After using these tool(s)/service(s), the author(s) reviewed and edited the content as needed and take(s) full responsibility for the publication’s content. 

\bibliography{sample-ceur}

\end{document}